\documentclass[letterpaper, 10 pt, conference]{ieeeconf}  

\IEEEoverridecommandlockouts                              

\usepackage{graphics}
\usepackage[pdftex]{graphicx}
\usepackage[tight,footnotesize]{subfigure}
\usepackage{multirow}
\usepackage[cmex10]{amsmath}
\usepackage{amssymb}
\usepackage{amsfonts}
\usepackage{algpseudocode}
\usepackage{algorithm}
\usepackage{booktabs}
\usepackage{arydshln}
\usepackage{textcomp}
\usepackage{url}
\usepackage{color}
\usepackage[dvipsnames]{xcolor}
\usepackage{siunitx}
\usepackage{soul}
\usepackage{xcolor}
\usepackage{colortbl}
\usepackage{wrapfig}
\usepackage{babel}
\usepackage[small]{caption}

\makeatletter
\newcommand{\cdashmidrule}[1]{%
  \noalign{\vskip\aboverulesep}
  \cdashline{#1}
  \noalign{\vskip\belowrulesep}}
\makeatother

\usepackage[bookmarks=true,colorlinks]{hyperref}
\title{\LARGE \bf
DynamicGSG: Dynamic 3D Gaussian Scene Graphs \\ for Environment Adaptation
}

\author{Luzhou Ge$^{1\ast}$, Xiangyu Zhu$^{1\ast}$, Zhuo Yang$^{1}$ and Xuesong Li$^{1, \dagger}$
\thanks{$^{\ast}$ Equal contribution.}
\thanks{$^{\dagger}$ The corresponding author:{ \tt\small lixuesong@bit.edu.cn}}
\thanks{$^{1}$ School of Computer Science, Beijing Institute of Technology, China.}%
}

\begin{document}

\maketitle
\thispagestyle{empty}
\pagestyle{empty}

\begin{abstract}


In real-world scenarios, environment changes caused by human or agent activities make it extremely challenging for robots to perform various long-term tasks. Recent works typically struggle to effectively understand and adapt to dynamic environments due to the inability to update their environment representations in memory according to environment changes and lack of fine-grained reconstruction of the environments. To address these challenges, we propose \textit{DynamicGSG}, a dynamic, high-fidelity, open-vocabulary scene graph construction system leveraging Gaussian splatting. \textit{DynamicGSG} builds hierarchical scene graphs using advanced vision language models to represent the spatial and semantic relationships between objects in the environments, utilizes a joint feature loss we designed to supervise Gaussian instance grouping while optimizing the Gaussian maps, and locally updates the Gaussian scene graphs according to real environment changes for long-term environment adaptation. Experiments and ablation studies demonstrate the performance and efficacy of our proposed method in terms of semantic segmentation, language-guided object retrieval, and reconstruction quality. Furthermore, we validate the dynamic updating capabilities of our system in real laboratory environments. The source code and supplementary experimental materials will be released at:~\href{https://github.com/GeLuzhou/Dynamic-GSG}{https://github.com/GeLuzhou/Dynamic-GSG}.
\end{abstract}

\section{Introduction}  \label{sec:intro}

Future intelligent robots are supposed to execute diverse long-term complex tasks in dynamic environments based on intricate natural language instructions from humans. To achieve this, agents must possess dynamic environment perception and comprehension capabilities. Prior studies \cite{gu2023conceptgraphsopenvocabulary3dscene, werby23hovsg, yan2025dynamicopenvocabulary3dscene, linok2024barequeriesopenvocabularyobject} construct static open-vocabulary scene graphs from sensor data to capture the topological structures of environments at specific moments, which enhances robots' understanding of complex instructions and facilitates task completion. However, static scene graphs are of limited use in real-world scenarios, as robot workspaces typically change due to human activities or other agents' operations. Additionally, the inherent latency between the scene graph stored in memory (which guides task planning and execution) and the actual environment state significantly hinders successful task completion.

Recent advancements in 3D Gaussian splatting~\cite{kerbl20233d, Huang2DGS2024,yu2024mip,cheng2024gaussianpro} have attracted significant attention within the robotics community. These developments have applications in various domains, including high-quality reconstruction~\cite{kerbl20233d, Huang2DGS2024, keetha2024splatam, yan2024gs} robotic manipulation~\cite{zheng2024gaussiangrasper3dlanguagegaussian}, semantic embedding~\cite{ye2024gaussiangroupingsegmentedit, Li_2024}, and 3D environment understanding~\cite{qin2024langsplat3dlanguagegaussian,wu2024opengaussianpointlevel3dgaussianbased}. The explicit point-based representation of 3D Gaussians effectively facilitates semantic information integration from advanced vision language models, enabling the construction of topological scene representations. Based on fast differentiable rendering and explicit representation, we find 3D Gaussians are particularly suitable for locally rapid updates of reconstructed scenes. Moreover, the high-fidelity environment reconstruction provided by 3D Gaussians naturally supports the development of dynamic, high-quality scene graph construction systems.

\begin{figure}
  \centering
  \includegraphics[width=1.0\linewidth]{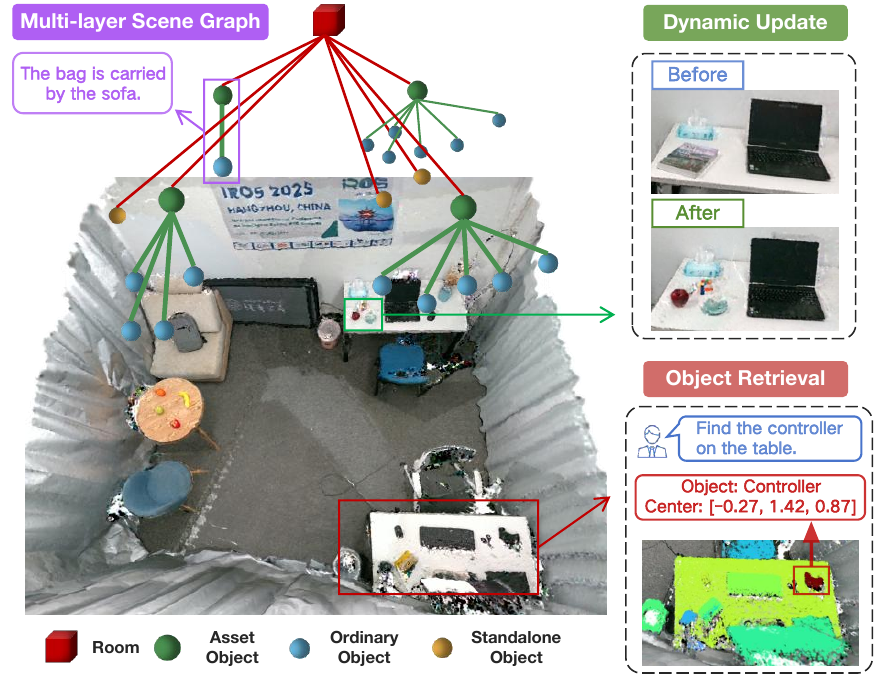}
  \caption{The dynamic high-fidelity multi-layer Gaussian scene graphs we constructed can adapt to environment changes, represent the spatial and semantic relationships of the objects, and support various forms of language-guided object retrieval.}
  \label{fig:teaser}
\end{figure}

Most previous works on scene graph construction primarily rely on point clouds~\cite{gu2023conceptgraphsopenvocabulary3dscene, werby23hovsg, linok2024barequeriesopenvocabularyobject, hughes2022hydra, hughes2024foundations} These methods often struggle to promptly respond to dynamic environment changes and fail to capture fine-grained details of 3D scenes due to the inherent limitations of traditional representation techniques. 

In this paper, we propose \textbf{DynamicGSG}, a framework that utilizes 3D Gaussian Splatting to construct dynamic, high-fidelity, open-vocabulary scene graphs. Advanced vision models such as Yolo-World \cite{cheng2024yoloworldrealtimeopenvocabularyobject}, Segment Anything \cite{kirillov2023segment}, CLIP \cite{radford2021learningtransferablevisualmodels} are employed to detect objects and extract their semantic features. Subsequently, we analyze the spatial and semantic relationships of objects to build hierarchical scene graphs using Large Vision Language Model (LVLM). Through incorporating additional semantic supervision, we improve the accuracy of instance-level Gaussian grouping and the overall reconstruction quality. Moreover, the rapid training and differentiable rendering of 3D Gaussians facilitate efficient scene updates to accommodate environment changes.  

In summary, our contributions are as follows:
\begin{itemize}
    \item We propose \textbf{Dynamic} 3D \textbf{G}aussian \textbf{S}cene \textbf{G}raphs, combining instance-level rendering with VLM semantic information to achieve 3D-2D object association and building multi-layer scene graphs with LVLM.
    \item We design a joint loss function that ensures accurate intra-instance Gaussian grouping and high-fidelity scene reconstruction.
    \item We utilize the fast differentiable rendering of Gaussians to update the scene graphs, enabling our system to adapt to dynamic environment changes.
    \item We deployed DynamicGSG in real-lab environments, demonstrating its capability to construct 3D Gaussian scene graphs and perform dynamic updates for effective environment adaptation.
\end{itemize}

\section{Related Work}  \label{sec:related_work}
\subsection{3D Scene Graphs}
\label{sec:3d_scene_graph}
3D scene graph \cite{rosinol20203ddynamicscenegraphs,hughes2024foundations} offers a hierarchical graph-structured representation of the environment with nodes representing spatial concepts at multiple levels of abstraction (e.g., objects, places, rooms, buildings.) and edges preserving the spatial and semantic relationships between nodes. Taking advantage of the generalization abilities of vision foundation models \cite{cheng2024yoloworldrealtimeopenvocabularyobject,kirillov2023segment, liu2024groundingdinomarryingdino, zhang2023recognizeanythingstrongimage} and cross-modal grounding capabilities of vision-language models \cite{radford2021learningtransferablevisualmodels, zhai2023sigmoidlosslanguageimage} enables the construction of 3D scene graphs at the open-vocabulary level. Recent methods \cite{gu2023conceptgraphsopenvocabulary3dscene, werby23hovsg, linok2024barequeriesopenvocabularyobject, yan2025dynamicopenvocabulary3dscene, jiang2024roboexpactionconditionedscenegraph} construct 3D scene graphs embedded with VLM semantic features, facilitating object retrieval, robot manipulation and navigation. 
 
However, most of these methods \cite{gu2023conceptgraphsopenvocabulary3dscene, werby23hovsg,linok2024barequeriesopenvocabularyobject} rely on static environment assumptions and typically lack mechanisms for handling dynamic updates. RoboEXP \cite{jiang2024roboexpactionconditionedscenegraph} conducts dynamic scene updates based on object spatial relationships and the scene modifications resulting from robot operations. DovSG \cite{yan2025dynamicopenvocabulary3dscene} extends dynamic scene graphs construction to mobile agents and utilizes Large Language Models (LLMs) \cite{openai2024gpt4technicalreport} for task decomposition and planning, enabling robots to accomplish complex tasks in dynamic environments over the long term. To enhance computational efficiency, these methods often employ aggressive point cloud downsampling strategies, which significantly limits their capacity for high-fidelity geometric reconstruction of scene details.

\subsection{Gaussian-based Open-Vocabulary Scene Understanding}
\label{sec:gaussian_open_vocab_understand}
Recent progress in Gaussian Splatting~\cite{kerbl20233d, Huang2DGS2024, yu2024mip, guedon2024sugar, cheng2024gaussianpro} demonstrates outstanding performance in photo-realistic reconstruction with promising efficiency. The gradient-based optimization has also been applied in an online setting to incrementally construct the Gaussian map \cite{keetha2024splatam,yan2024gs, matsuki2024gaussian, wei2024gsfusiononlinergbdmapping} through differentiable rendering.

Concurrently, advancements in vision foundation models (such as SAM \cite{kirillov2023segment}, CLIP \cite{radford2021learningtransferablevisualmodels}, DINO \cite{oquab2024dinov2learningrobustvisual}) have motivated the exploration of integrating 2D semantic features into 3D Gaussians. LangSplat \cite{qin2024langsplat3dlanguagegaussian} learns hierarchical semantics using SAM and trains an autoencoder to distill high-dimensional CLIP features into low-dimensional semantic attributes of Gaussians.  GaussianGrouping \cite{ye2024gaussiangroupingsegmentedit} implements joint 3D Gaussian optimization based on 2D pre-matched SAM masks and 3D spatial consistency,  enabling high-quality scene reconstruction and open-vocabulary object segmentation. OpenGaussian \cite{wu2024opengaussianpointlevel3dgaussianbased} augments 3D Gaussian Splatting with point-level open-vocabulary understanding through SAM-based instance feature training, a two-level codebook for discretization, and instance-level 3D-2D feature association for geometric-semantic alignment. However,  Most of the above works require embedding semantic information into pre-trained Gaussian scenes or conducting data preprocess prior to gradient-based optimization. Such offline pipelines inherently conflict with robots' operational paradigms in dynamic settings.


To address these deficiencies, we incrementally construct the semantic Gaussian map using posed RGB-D sequences from public datasets or live cameras running Visual-Inertial Odometry (VIO) \cite{qin2017vins, qin2018online, qin2019a, qin2019b} frameworks, enabling it to achieve high-fidelity reconstruction and integrate rich semantic information to build a topological scene graph. Leveraging the rapid training and differentiable rendering capabilities of 3D Gaussians, our system can dynamically update both the Gaussian maps and scene graphs to adapt to changes in the real-world environments.

\begin{figure*}
  \centering
  \includegraphics[width=1.0\textwidth]{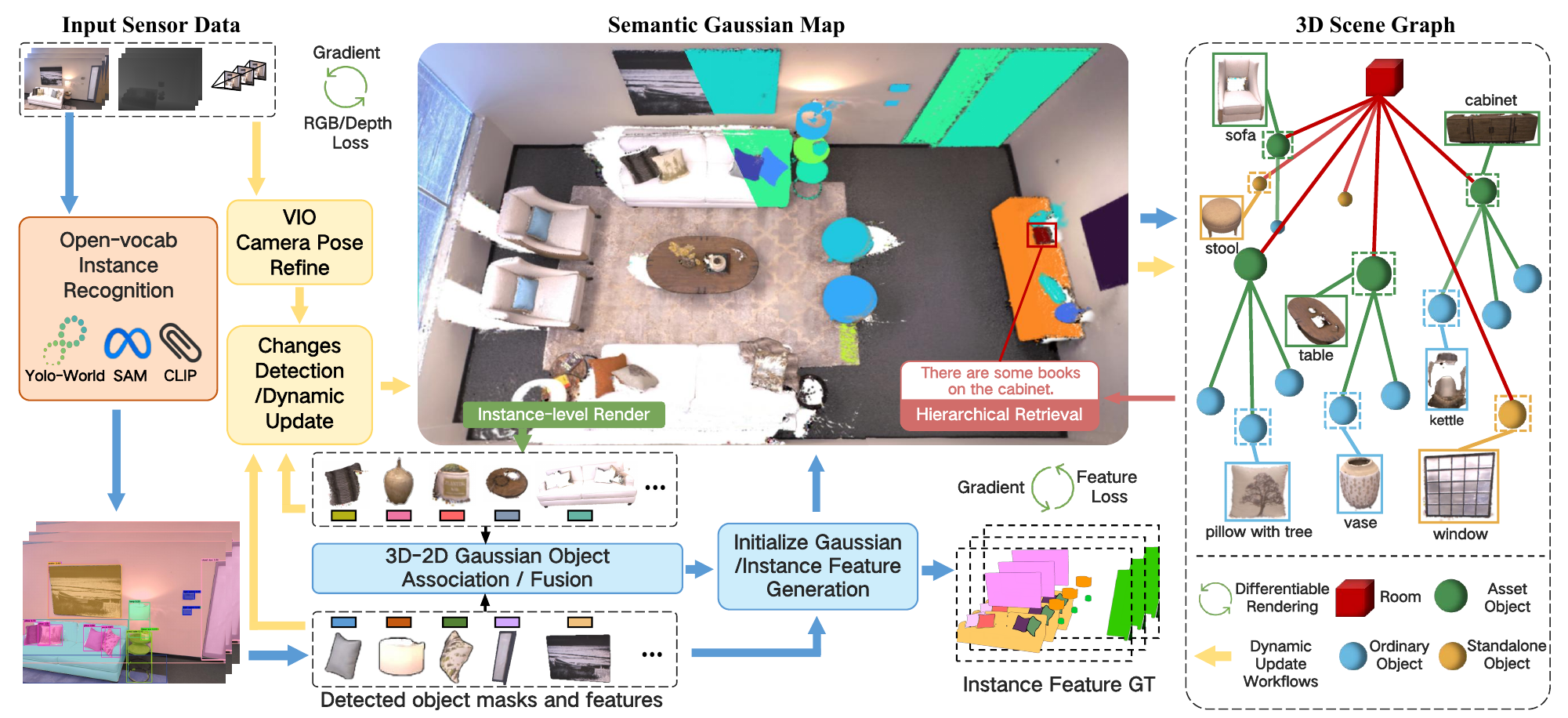}
  \caption{\textbf{Overview of DynamicGSG}: Our system processes posed RGB-D sequences, utilizes open-vocabulary object detection and segmentation models to obtain 2D masks, and extracts corresponding semantic features. In parallel, we employ instance-level rendering to get 2D masks and semantic features of objects in the map for object fusion. Subsequently, we perform Gaussian initialization and joint optimization to incrementally create a high-fidelity object-centric Gaussian map. Based on the spatial relationship of objects and the capabilities of LVLM, we construct a hierarchical scene graph to provide a structured description of the scene. In dynamic real-world scenarios, after refining the initial camera poses obtained from VINS-Fusion\cite{qin2019a}, we detect local changes and make corresponding modifications in the Gaussian map and scene graph for environment adaptation.
}
  \label{fig: pipeline}
\end{figure*}

\section{Methodology}
\label{sec:methodology}
The proposed system constructs dynamic high-fidelity Gaussian scene graphs by splatting Gaussians onto 2D image planes in various forms, building semantic scene graphs, optimizing Gaussian maps, and making dynamic updates. An overview of DynamicGSG is presented in Fig.~\ref{fig: pipeline}.

\subsection{Gaussian Preliminaries}
\label{subsec:gaussians}
We densely represent the scene using isotropic Gaussian, which is an explicit representation parameterized by RGB color $\mathbf{c}$, center position $\boldsymbol{\mu} \in \mathbb{R}^3 $, radius $\mathbf{r}  \in \mathbb{R}^+$, and opacity $\mathbf{o} \in [0,1]$. The influence of each Gaussian on 3D space point $\mathbf{x} \in \mathbb{R}^3$ is defined as:
\begin{equation}
\label{eq:opacity}
f(\mathbf{x}) = \mathbf{o} \cdot \exp\left( \frac{\|\mathbf{x} - \boldsymbol{\mu}\|^2}{2\mathbf{r}^2}\right)
\end{equation}

The view synthesis and Gaussian parameters optimization are implemented through differentiable rendering with the Gaussian map and a camera pose $T \in SE(3)$. The color, depth, and visibility (accumulated opacity) of each pixel $\mathbf{u}$ at camera pose $T_t$ is determined by $\alpha$-blending contributions from depth-ordered projections of 3D Gaussians:
\begin{equation}
\label{eq:render_rgb}
\hat{C}_t(\mathbf{u})=\sum_{i=1}^n\mathbf{c}_if_i(\mathbf{u})\prod_{j=1}^{i-1}\left(1-f_j(\mathbf{u})\right).
\end{equation}
\begin{equation}
\label{eq:render_depth}
    \hat{D}_t(\mathbf{u}) =\sum_{i=1}^n d_if_i(\mathbf{u})\prod_{j=1}^{i-1}\left(1-f_j(\mathbf{u})\right),
\end{equation}
\begin{equation}
\label{eq:render_opacity}
\hat{S}_t(\mathbf{u}) =\sum_{i=1}^nf_i(\mathbf{u})\prod_{j=1}^{i-1}\left(1-f_j(\mathbf{u})\right),
\end{equation}
where $d_i$ is the depth of the i-th Gaussian center in the camera coordinate.

We preserve the basic properties of Gaussians while introducing two additional parameters: a low-dimensional instance feature $\mathbf{e} \in \mathbb{R}^3$ and an identifier $\mathbf{idx} \in \mathbb{N}^+$. We render the 2D instance feature $\hat{E} \in \mathbb{R}^{3\times H\times W}$ for each pixel in a differentiable manner similar to color blending:
\begin{equation}
\label{eq:render_feature}
\hat{E}_t(\mathbf{u})=\sum_{i=1}^n\mathbf{e}_if_i(\mathbf{u})\prod_{j=1}^{i-1}\left(1-f_j(\mathbf{u})\right).
\end{equation}

\subsection{Gaussian Objects Association}
\label{subsec:obj_association}
Our system processes posed RGB-D sequences $ I_t= \left \langle C_t, D_t, T_t \right \rangle  \in \{ I_1, I_2, \dots, I_t \} $ to construct a Gaussian scene graph: $\textbf{G}=\{O_{t}, E_t\}$, where $ O_t=\{ o_j \}_{j=1}^{J} $ and $ E_t=\{ e_k \}_{k=1}^{K} $ represent the sets of objects and edges, respectively. Each object $o_j$ is characterized by a set of Gaussians indexed by $ idx $ and a semantic feature vector $ f_{o_j} $.

\paragraph{Object Recognition}
\label{para:obj_recogniton}
We employ various advanced vision models to obtain instance-level semantic information from RGB images. For the current frame $I_t$, we first use an open-vocabulary object detection model (YOLO-World \cite{cheng2024yoloworldrealtimeopenvocabularyobject}) to obtain object bounding boxes $ \{\textbf{b}_{t,i}\}_{i=1}^{M}$. These detected proposals are subsequently processed by Segment Anything \cite{kirillov2023segment} to generate corresponding masks $ \{\textbf{m}_{t,i}\}_{i=1}^{M}$. Post-processing is then performed to ensure these masks do not overlap with each other. Finally, we use CLIP \cite{radford2021learningtransferablevisualmodels} to obtain each detection's semantic feature. Since CLIP's visual descriptors are image-aligned, each mask $\textbf{m}_{t,i}$ is passed to CLIP to extract an instance-level semantic feature $f_{t,i}$.

\paragraph{3D-2D Object Association}
\label{para:obj_association}

\begin{figure}[t]
  \centering
  \includegraphics[width=1.0\linewidth]{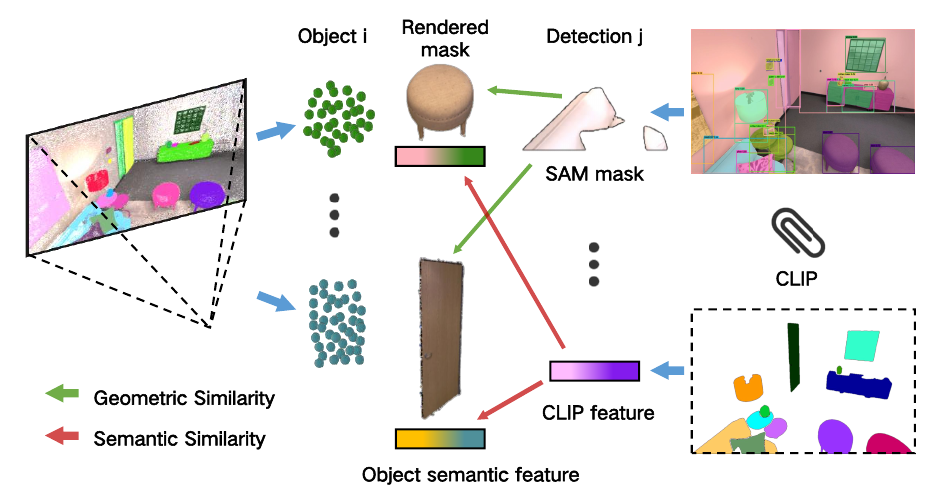}
  \caption{\textbf{3D-2D Gaussian Object Association.}}
  \label{fig:Association}
\end{figure}

For all detections $ D_t = \{ \textbf{d}_{t,i} \}_{i=1}^{M} $ in $ I_t $, we associate them with the map objects $ O_{t-1} $  through joint geometric and semantic similarity matching. As shown in 
we first render all Gaussians into the current camera view at instance-level to obtain object masks  $\{\textbf{m}_{t,o_j}\}_{j=1}^{J}$. Objects $O_{t-1}$ whose projected masks contain more than $ \delta_{\text{pix}} $ pixels in the image plane are considered visible in the current frame. The geometric similarity is then formulated by the Intersection over Union (IoU) between all 2D detection masks $ \{\textbf{m}_{t,i}\}_{i=1}^{M}$ and projected masks of visible 3D objects $\{\textbf{m}_{t,o_j}\}_{j=1}^{J}$:
\begin{equation}
s_{\text{geo}}(i, j) = \frac{m_{t,i} \cap m_{t,o_j}}{m_{t,i} \cup m_{t,o_j}}
\end{equation}

The semantic similarity is calculated as the normalized cosine similarity between detection CLIP descriptors $\{f_{t, i}\}_{i=1}^{M}$ and visible object semantic features $\{f_{o_j}\}_{j=1}^{J}$. The joint similarity $ s(i,j) $ combines geometric and semantic similarity through weighted summation:
\begin{equation}
s_{\text{sem}}(i,j) = (f_{t,i})^\top f_{o_j} /2+1/2
\end{equation}
\begin{equation}
s(i,j) = w_g s_{\text{geo}}(i,j) + w_s s_{\text{sem}}(i,j)
\end{equation}
where $ w_g + w_s=1 $. Object association is performed through a greedy algorithm that assigns each detection to a visible object with the maximum similarity score. A new object is initialized when no visible object with similarity exceeding the threshold $\delta_{sim}$ is matched.

\paragraph{Object Fusion}
\label{para:obj_fusion}

If a detection $d_{t,i}$ is associated with a map object $o_j$, we fuse this detection with the corresponding object by updating the object’s semantic feature as:
\begin{equation}
f_{o_j}=(n_{o_j}f_{o_j}+f_{t,i})/(n_{o_j}+1)
\end{equation}
where $n_{o_j}$ denotes the number of $o_j$ that has been associated.

\subsection{Gaussian Map Optimization}
\label{subsec:gaussian_optim}
To incrementally construct the object-centric Gaussian map, new Gaussians must be initialized and subsequently optimized with observations to ensure our map is realistic and accurately group Gaussians belonging to each object.

\paragraph{Densification based on opacity}
\label{para:densification}

Firstly, new Gaussians will be dynamically initialized to cover newly observed regions and objects. We create a newly observed mask following~\cite{keetha2024splatam}, which identifies pixels exhibiting either insufficient transparency accumulation or emerging geometry occluding the existing scene structure: 
\begin{equation}
M_t = (\hat{S}_t < \lambda_s) \cup  ((D_t < \hat{D}_t) (|D_t - \hat{D}_t|) > \lambda_\text{MDE} )
\end{equation}
where $\lambda_s = 0.5$, $\lambda_{MDE}$ equals 50 times median depth error. For each pixel in $I_t$, we add a new Gaussian characterized by the pixel's color, aligned depth, and an opacity of 0.5. Each new Gaussian's $idx$ is assigned by the object's index after object association, and its instance feature $e$ is initialized using the value corresponding to $idx$ in the codebook, which contains 200 instance colors from ScanNet dataset~\cite{dai2017scannet}. As illustrated in the lower-right corner of Fig. \ref{fig: pipeline},
these instance features are then projected onto the 2D mask regions where the corresponding objects is located in $C_{t}$, forming a ground-truth instance feature $E_t  \in \mathbb{R}^{3\times H\times W}$.

\paragraph{Intra-instance Gaussian Regularization}
\label{para:regularization}

During gradient-based optimization, each attribute of Gaussians will be refined according to differentiable rendering. As shown in Fig. \ref{fig:feature_ablation},
we observe that Gaussians initially grouped by object indexes tend to progressively encroach into regions occupied by other objects, resulting in segmentation artifacts and reduced boundary precision. To address this limitation, we propose a novel feature consistency loss with GT instance feature $E_t$ to improve reconstruction fidelity and instance-level grouping accuracy:
\begin{equation}
L_\text{feature} = \lambda_1|E_t - \hat{E}_t| + \lambda_2(1-SSIM(E_t, \hat{E_t}))
\end{equation}
where $\lambda_1 = 0.8, \lambda_2 = 0.2$. As demonstrated in Fig. \ref{fig:feature_ablation}, our feature loss enforces intra-instance Gaussian consistency while boosting semantic segmentation precision.

\paragraph{Gaussian map optimization}
\label{para:map_optimization}
We minimize the joint loss, including color, depth, and features, to optimize the Gaussian map:
\begin{equation}
L_\text{color} = \lambda_3|C_t - \hat{C}_t| + \lambda_4(1-SSIM(C_t, \hat{C}_t))
\end{equation}
\begin{equation}
L_\text{depth} = |D_t - \hat{D}_t|
\end{equation}
\begin{equation}
L_\text{mapping} = w_c L_\text{color} + w_d L_\text{depth} + w_f L_\text{feature}
\end{equation}
where $\lambda_3 = 0.8, \lambda_4 = 0.2, w_c = w_f =0.5, w_d = 1.0$.

In the process of optimizing the Gaussian map, we perform multi-view scene optimization by collecting a list of keyframes to improve the quality of 3D reconstruction. A keyframe is selected and stored every $n$-th frame, and $m$ keyframes are chosen for multi-view optimization based on temporal distance and geometric constraints. Furthermore, we prune redundant Gaussians with near-zero opacity or large covariances as \cite{kerbl20233d}.

\subsection{Multi-layer Scene Graph Construction}
\label{subsec:scene_graph_generation}
With the well-reconstructed object-centric Gaussian map, we construct a hierarchical scene graph $\textbf{G}_t =\{O_{t}, E_t\}$ that reflects the structured description of the environment. As illustrated on the right side of Fig.~\ref{fig: pipeline}, objects are categorized into asset, ordinary, and standalone objects.

\paragraph{Asset objects}
\label{para:asset_layer}
We use a Large Vision Language Model (GPT-4o) to identify asset objects in indoor environments. The expected asset objects typically include furniture on the ground, such as chairs, tables, or cabinets, rather than small portable containers or decorative items (e.g., baskets, vases, or lamps). Our system takes the labels generated by Yolo-World\cite{cheng2024yoloworldrealtimeopenvocabularyobject} for objects in $O_t$ as input to GPT-4o. The asset objects, denoted as $ \tilde{O}_t  \subseteq O_t$, are classified through a specific prompt.


\paragraph{Ordinary objects}
\label{para:ordinary_layer}
We define ordinary objects $\Bar{O}_t$ through spatial relationships between an object and asset objects $\tilde{O}_{t}$. If the center position of an object is located above the spatial range of the asset object $\tilde{o}_{j} \in \tilde{O}_{t}$, we determine this object as $\Bar{o}_{i} \in \Bar{O}_t$ and  establish an edge $e_t(i,j) \in E_t$ between them. This edge not only represents the spatial relationship but also indicates that $\Bar{o}_{i}$ is carried by $\tilde{o}_{j}$.

\paragraph{Standalone objects}
\label{para:standalone_layer}
Standalone objects are defined as the complement of asset and ordinary objects, represented as $\hat{O}_t \subseteq (O_t-\tilde{O}_t - \Bar{O}_t)$. Typical instances include stools, windows, and similar entities that exhibit inherent independence in operational environments.

Additionally, asset and standalone objects are directly connected to the room node within the scene graph $\textbf{G}_t$.

\subsection{Dynamic Scene Update}
\label{para:dynamic_update}

In dynamic human-robot collaborative environments, the changes in object positions and their spatial interrelationships present significant challenges for long-term task execution. Leveraging high-fidelity reconstruction and fast differentiable rendering of 3D Gaussian, we implement a dynamic update mechanism to detect changes and locally update the Gaussian map and scene graph at the instance level with real-time RGB-D observations, ensuring temporal consistency between the environment representations in the robot's memory and the physical workspace, as shown in Fig. \ref{fig:teaser} and \ref{fig:dynamic}.

\paragraph{Refine Camera Tracking}
\label{para:camera_tracking}
In public datasets or simulation environments, our system can directly utilize RGB-D frames with ground-truth poses as input. In real-world scenarios, the lack of reliable pose priors introduces fundamental constraints on the construction and modification of Gaussian scene graphs. To address this sensing gap, we deploy a validated and efficient VIO\cite{qin2017vins} framework to acquire an initial estimate $T_{t,est}$ and execute an iterative pose refinement by minimizing the L1 loss between aligned color $C_{t}$ and depth $D_{t}$ from the camera and their rendered views to obtain precise camera pose $T_{t}$:
\begin{equation}
    L_\text{tracking} = (\hat{S}_t > \lambda_r)(\lambda_5|C_t - \hat{C}_t| + \lambda_6|D_t - \hat{D}_t|)
\end{equation}
where $\lambda_r = 0.99, \lambda_5 = 0.5, \lambda_6 = 1.0$. During this stage, the parameters of the Gaussians are fixed.

\paragraph{Update Local Gaussian Map}
\label{para:update_map}
Using the refined pose $T_{t}$, our system begins to detect local changes, including the disappearance, displacement, and emergence of objects. Subsequently, we will modify the map and alter the scene graph at the instance level.

To address the disappearance and movement of objects, we render all objects where over 50\% of their Gaussians appear within the current camera frustum to acquire visible object RGB masks $\{m_{t, o_j}^{rgb}\}_{j=1}^{J}$ and compute structural similarity (SSIM) between $\{m_{t, o_j}^{rgb}\}_{j=1}^{J}$ and the corresponding regions within RGB observation $C_t$:
\begin{equation}
S(j) = SSIM(m_{t, o_j}^{rgb}, C_t(m_{t, o_j}))
\end{equation}
If $ S(j) < \delta_{\text{change}} $, it indicates that object $o_j$ has either been moved or substituted by another object. We categorize all disappeared objects as $O_\text{delete}$ and remove their associated obsolete Gaussians. Next, new objects $O_\text{appear}$ will be instantiated following Sec. \ref{subsec:obj_association}, based on the recalculated similarity matrix $s(i, j)$ between $\{\textbf{d}_{t,i}\}_{i=1}^{M}$ and $\{\textbf{o}_{t-1,j}\}_{j=1}^{J'} \subseteq \{O_{t-1} - O_\text{delete}\}$. The union of deleted and new objects constitutes the update set $O_\text{update} = \{O_\text{appear}, O_\text{delete}\}$.

Finally, if $O_\text{update} \neq \varnothing$, our system will clear the keyframe list to prevent outdated keyframes from participating in the subsequent Gaussian map optimization detailed in Sec. \ref{subsec:gaussian_optim}. Otherwise, we shall skip this stage.

\paragraph{Update Scene graph}
\label{para:update_scene_graph} 

Following the local updates of the Gaussian map, the scene graph $\mathbf{G}_{t-1}$ requires corresponding adjustments based on the categories of objects in $O_\text{update}$. For the deleted objects $O_\text{delete}$, if an object is marked as ordinary or standalone, it is sufficient to delete it along with the parent edge $e_k$ from $\textbf{G}_{t-1}$; in contrast, if the object is an asset object, the removal must extend to all its child nodes and their associated edges. For the newly appeared objects $O_\text{appear}$, we determine their type using the method described in Sec.~\ref{subsec:scene_graph_generation}. Subsequently, we insert nodes of corresponding types and establish relational edges within $\textbf{G}_{t-1}$ to construct the updated scene graph $\textbf{G}_t$.


\section{Experiment}
\label{sec:experiment}
\subsection{Experiment Setups}
\label{subsec:experiment_setup}

\renewcommand{\arraystretch}{1.2}
\begin{table}[!t]
    \centering
    \begin{tabular}{lccc}\toprule
\multirow{2}{*}{\textbf{Methods}}      & \multicolumn{3}{c}{\textbf{Metrics}} \\
& mAcc$\uparrow$        & mIoU$\uparrow$   & F-mIoU$\uparrow$   \\ \midrule 

ConceptGraphs~\cite{gu2023conceptgraphsopenvocabulary3dscene}       & 39.43                    & 25.57   & 44.06     \\

ConceptGraphs-Detector~\cite{gu2023conceptgraphsopenvocabulary3dscene} & 41.18                        & 26.82       & 42.28             \\

HOV-SG~\cite{werby23hovsg}  & 39.95                     & 27.52        & \textbf{46.79}      \\

\textbf{DynamicGSG}   & \textbf{54.04}              & \textbf{31.06}               & 46.21               \\

\textbf{DynamicGSG} w/o feature loss   & 52.94                                & 25.97     & 37.32       \\
 \bottomrule
\end{tabular}
    \caption{\textbf{3D Open-vocabulary Semantic Segmentation on Replica~\cite{replica19arxiv}:} Attributed to 3D-2D Gaussian Object Association, DynamicGSG significantly outperforms the baselines in terms of both mAcc and mIOU. And the joint feature loss also effectively improve the mIOU and F-mIOU of semantic segmentation.}
    \label{tab:semseg}
    \vspace{-1em}
\end{table}

\begin{figure}[tb]
  \centering
  \includegraphics[width=1.0\linewidth]{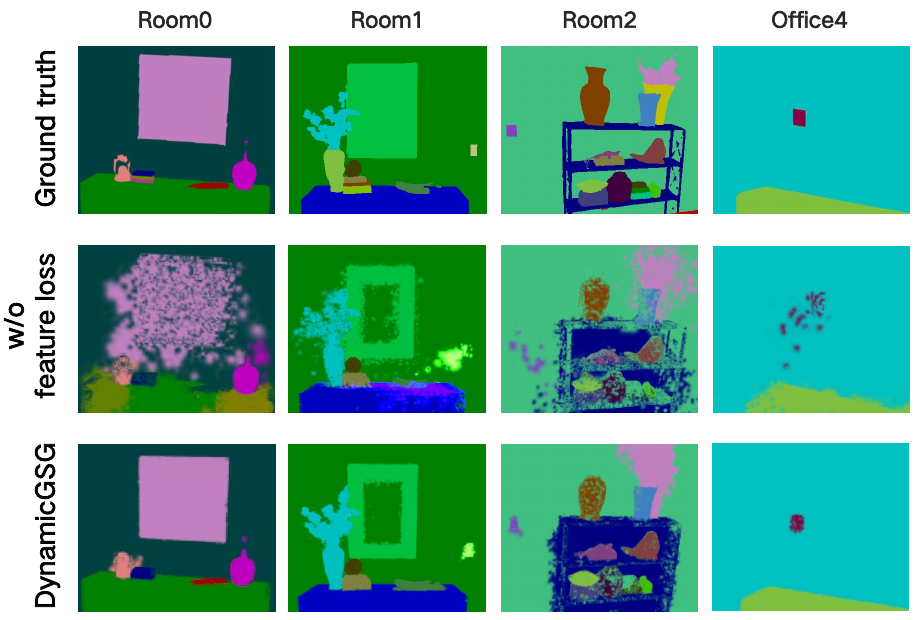}
  \caption{\textbf{Visualization of Feature Loss Ablation Experiments.}}
  \label{fig:feature_ablation}
      \vspace{-1em}
\end{figure}

To comprehensively evaluate DynamicGSG, we conduct a series of experiments using data sourced from Replica\cite{replica19arxiv}, ScanNet++\cite{yeshwanthliu2023scannetpp}, and real laboratory environments: (1) A quantitative comparison of 3D open-vocabulary semantic segmentation on the Replica dataset, contrasting our results with recent open-vocabulary scene graph construction methods, accompanied by an ablation study to investigate the contribution of joint feature loss. (2) A language-guided object retrieval experiment to evaluate the effectiveness of multi-layer scene graphs generated by DynamicGSG in capturing spatial-semantic object relationships. (3) Quantitative evaluation of scene reconstruction quality on Replica and ScanNet++ datasets. (4) Within our laboratory, we manually introduce environment changes to validate DynamicGSG's capability for dynamic updating of Gaussian scene graphs. 

All experiments are conducted on a desktop computer equipped with an Intel Core i7-14700KF CPU, an NVIDIA RTX 4090D GPU, and 32GB RAM. In all experiments, we set thresholds $ \delta_{\text{pix}} $ = 200, $ \delta_{\text{sim}} $ = 0.55 and $ \delta_{\text{change}} $ = 0.15.

\renewcommand{\arraystretch}{0.9}
\begin{table}[!t]
    \centering
    \setlength{\tabcolsep}{5pt}  
\begin{tabular}{llccccc}
    \toprule
    \textbf{Methods} & \textbf{Query}  & \textbf{Match} & \textbf{R@1} & \textbf{R@2} & \textbf{R@3}  \\
    \midrule
    \multirow{8}{*}{ConceptGraphs~\cite{gu2023conceptgraphsopenvocabulary3dscene}} & \multirow{3}{*}{Descriptive} & CLIP &  0.52 &  0.65 & 0.71  \\
                             &                              &  LLM  &  0.40 &  0.55 &  0.62 & \\
                             &                              & 
                             HSG  &  -- &  --   &  --   & \\
    \cdashmidrule{2-6}
                             & \multirow{2}{*}{Affordance}  &  CLIP &  0.60 &  0.63 &  \textbf{0.69} \\
                             &                              &  LLM  &  0.60 &  0.69 &  \textbf{0.80}& \\
    \cdashmidrule{2-6}
                             & \multirow{2}{*}{Negation}    &  CLIP &  0.17 &  0.49 &  0.60 \\
                             &                              &  LLM  &  0.77 &  \textbf{0.91} &  \textbf{0.97} & \\

    \midrule
    \multirow{8}{*}{\textbf{DynamicGSG}}     & \multirow{3}{*}{Descriptive} &  CLIP &  \textbf{0.64} &  \textbf{0.74}   &  \textbf{0.76}   \\
                             &                              &  LLM  &  \textbf{0.41} &  \textbf{0.57}   &  \textbf{0.64}  & \\
                             &                              & 
                              HSG  &  \cellcolor{Green!25}\textbf{0.71} &  \cellcolor{Green!25}\textbf{0.81}   &  \cellcolor{Green!25}\textbf{0.82}   & \\
    \cdashmidrule{2-6}                             
                             & \multirow{2}{*}{Affordance}  &  CLIP &  \textbf{0.60} &  \textbf{0.66} &  0.66  \\
                             &                              &  LLM  &  \textbf{0.65} &  \textbf{0.74} &  0.77 & \\
    \cdashmidrule{2-6}
                             & \multirow{2}{*}{Negation}    &  CLIP &  \textbf{0.34} &  \textbf{0.54}   &  \textbf{0.71}    \\
                             &                              &  LLM  &  \textbf{0.77} &  0.89   &  0.94   & \\

    \bottomrule
\end{tabular}

    \caption{\textbf{Language-guided Object retrieval on Replica\cite{replica19arxiv}.} CLIP, LLM, and HSG refer to Semantic-based match, LLM-based match, and Hierarchical scene graph-based match, respectively.}
    \label{tab:objret}
        \vspace{-1em}
\end{table}

\begin{figure}[tb]
  \centering
  \includegraphics[width=1.0\linewidth]{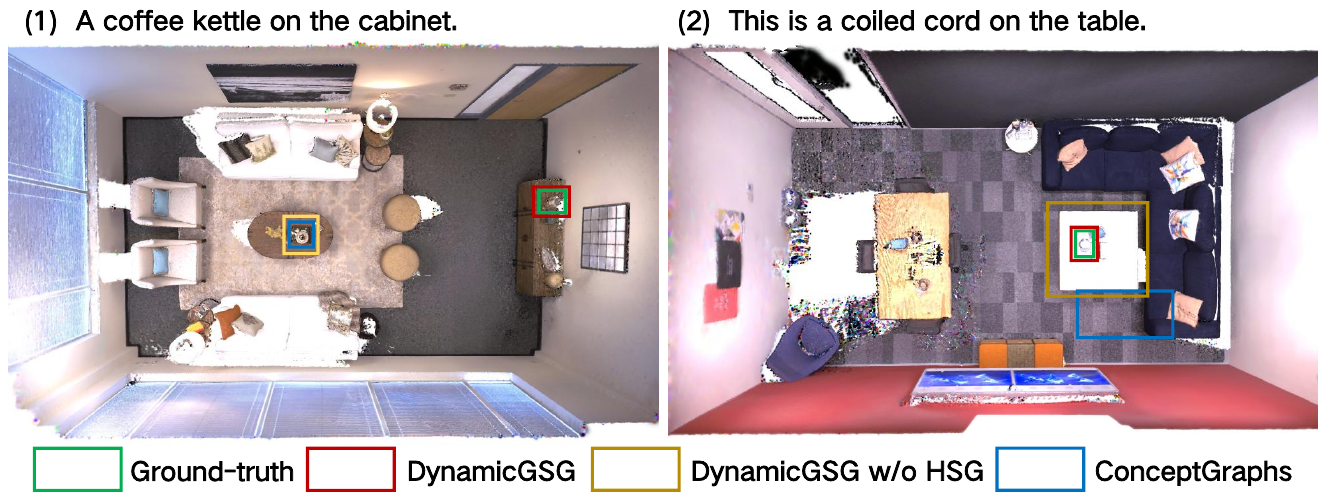}
  \caption{\textbf{Qualitative Results of Object Retrieval:} DynamicGSG effectively locates objects that ConceptGraphs~\cite{gu2023conceptgraphsopenvocabulary3dscene} cannot retrieve through Hierarchical scene graph-based match.}
  \label{fig:obj_retrieval}
  \vspace{-1em}
\end{figure}

\subsection{3D Open-vocabulary Semantic Segmentation}
\label{subsec:semantic_segmentation}
To evaluate the quality of semantic embeddings in DynamicGSG and investigate how joint feature loss supervision influences Gaussian instance grouping, we perform an ablation experiment of the 3D open-vocabulary semantic segmentation on 8 scenes from the Replica dataset \cite{replica19arxiv} and quantitatively compare our results with recent scene graph construction methods. The primary baseline methods used for comparison are ConceptGraphs \cite{gu2023conceptgraphsopenvocabulary3dscene} and HOV-SG \cite{werby23hovsg}. For the ablation analysis, we include a variant of DynamicGSG without feature loss. All compared methods consistently adopt the ViT-H-14 CLIP backbone for semantic feature extraction.

To generate the semantic segmentation, we first calculate the CLIP text description vector for class-specific prompt formatted as ``an image of \{class label\}" corresponding to each class in the Replica dataset. For each scene, we compute the cosine similarity between the semantic feature of each object within the scene graph and the text description vector of each class. Each object's points or Gaussians are allocated to the class with the highest similarity score. Finally, the point clouds or Gaussians generated by all methods are transformed to the same coordinate as the ground-truth semantic point clouds. Quantitative evaluation is performed through standardized metrics, including mAcc, mIoU, and frequency-weighted mIoU. 

As shown in Tab. \ref{tab:semseg}, our method performs better than all baselines on mAcc and mIoU while achieving comparable performance to HOV-SG on F-mIoU. Object association in \cite{gu2023conceptgraphsopenvocabulary3dscene, werby23hovsg} relies on the overlap ratio between point clouds which suffers from a critical limitation: the potential association of small objects into nearby large objects due to high overlap ratios of 3D point clouds. Our method employs 3D-2D object association which effectively prevents spurious merging, as the small masks and large objects do not exhibit abnormal geometric similarity, ultimately yielding a significant enhancement in mAcc. And the joint feature loss also effectively regularizes the Gaussian instance grouping to improve the mIOU and F-mIOU. The qualitative results in Fig. \ref{fig:feature_ablation} further demonstrate that joint feature loss significantly enhances the regularization of intra-instance Gaussian grouping.

\renewcommand{\arraystretch}{1.2}
\begin{table}[!t]
    \centering
    \begin{tabular}{lccc}\toprule
\multirow{2}{*}{\textbf{Methods}}      & \multicolumn{3}{c}{\textbf{Metrics}}  \\
& PSNR$\uparrow$        & SSIM$\uparrow$   & LPIPS$\downarrow$  \\ \midrule 

HOV-SG~\cite{werby23hovsg}      & 19.69          & 0.821     & 0.284  \\

ConceptGraphs~\cite{gu2023conceptgraphsopenvocabulary3dscene} & 23.24                & 0.910             & 0.204      \\

SGS-SLAM~\cite{Li_2024}  & 35.43            & 0.978      & 0.077       \\

\textbf{DynamicGSG}   & \textbf{35.62}       & \textbf{0.979}              & \textbf{0.068}              \\
 \bottomrule
\end{tabular}
    \caption{\textbf{Quantitative Reconstruction Performance on Replica~\cite{replica19arxiv}:} DynamicGSG is comparable to the Gaussian-based semantic SLAM method~\cite{Li_2024} while significantly outperforming methods based on point cloud~\cite{gu2023conceptgraphsopenvocabulary3dscene,werby23hovsg}.}
    \label{tab:replica}
\end{table}

\renewcommand{\arraystretch}{1.2}
\begin{table}[t]
\centering

\scriptsize
\setlength{\tabcolsep}{5.2pt}

\begin{tabular}{lcccccccc}
\toprule

\multirow{2}{*}{\textbf{Methods}} & \multirow{2}{*}{\textbf{Metrics}} & \multicolumn{3}{c}{\textbf{Train View}} & \multicolumn{3}{c}{\textbf{Novel View}} \\

 &   & \texttt{S1} & \texttt{S2} & \textbf{Avg.}  & \texttt{S1} & \texttt{S2} & \textbf{Avg.}\\

\midrule

\multirow{3}{*}{SplaTAM~\cite{keetha2024splatam}} & PSNR $\uparrow$   & 27.78 & 28.40 & 28.09  & 24.50 & 25.56 & 25.03\\
& SSIM $\uparrow$    & \textbf{0.946} & 0.944 & 0.945  & 0.896 & 0.892 & 0.894\\
& LPIPS $\downarrow$  & 0.121 & 0.129 & 0.125  & 0.210 & 0.255 & 0.233 \\

\cdashmidrule{1-8}

\multirow{3}{*}{\textbf{DynamicGSG}} & PSNR $\uparrow$  & \textbf{27.86} &  \textbf{28.47} &    \textbf{28.17}  & \textbf{24.81} & \textbf{25.65} & \textbf{25.19}\\
& SSIM $\uparrow$    &  0.945 &  \textbf{0.946} &  \textbf{0.946}  &  \textbf{0.902} &  \textbf{0.893} &  \textbf{0.898}\\
& LPIPS $\downarrow$   &  \textbf{0.120} &  \textbf{0.125} &  \textbf{0.123}  &  \textbf{0.202} &  \textbf{0.247} &  \textbf{0.225}\\

\bottomrule
\end{tabular}
\caption{\textbf{Novel \& Train View Synthesis Performance on ScanNet++~\cite{yeshwanthliu2023scannetpp}:} DynamicGSG not only provides photorealistic reconstruction on training views but also enables high-fidelity novel view synthesis at any camera pose.}
\label{tab:scannetpp}
      \vspace{-1em}
\end{table}

\subsection{Language-guided Object Retrieval}
\label{subsec:object_retrieval}

\begin{figure*}[!t]
  \centering
  \includegraphics[width=0.95\linewidth]{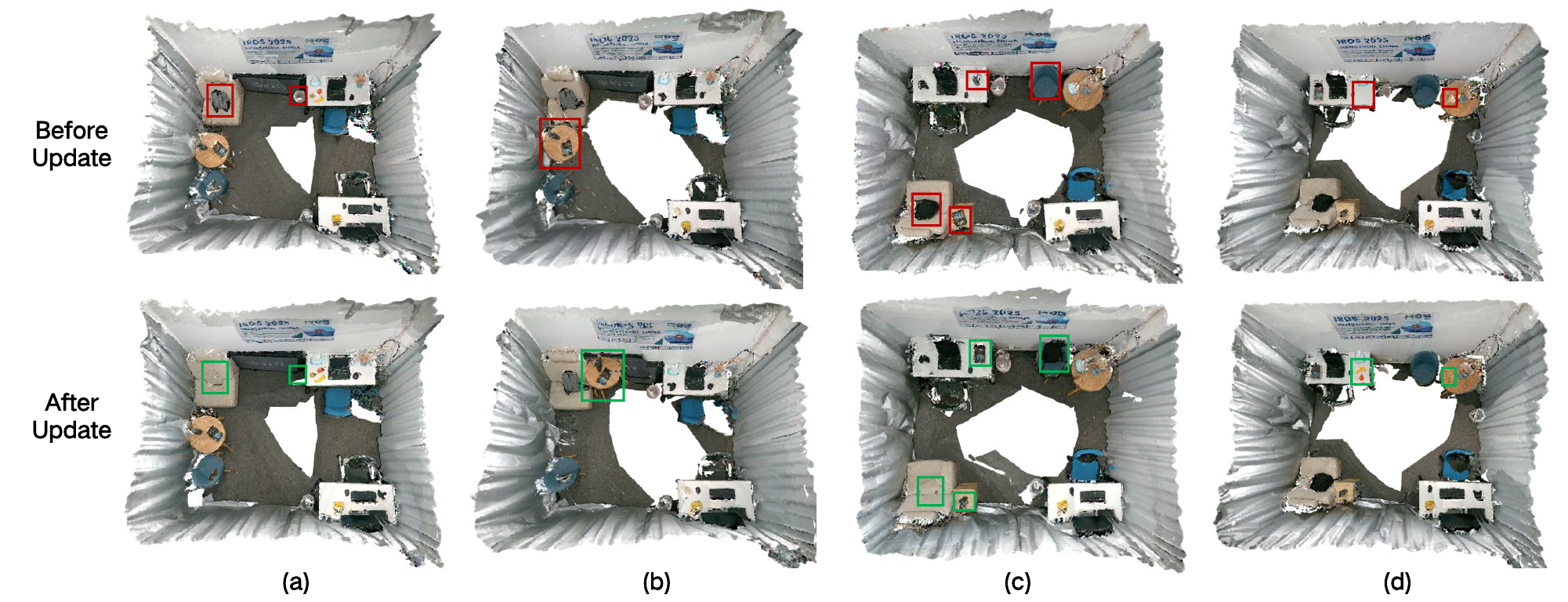}
  \caption{\textbf{Visualization of Dynamic Updates}: (a) The backpack on the sofa and the bin are removed. (b) Holistic position update of the table and its contents (books and bottle). (c) The books and bottle exchange positions, and the backpack is moved to the chair. (d) The teacup disappears from the tea table and some fruits appear on the computer table.}
  \label{fig:dynamic}
\end{figure*}

To validate whether the multi-layer scene graphs constructed by DynamicGSG can effectively capture spatial and semantic object relationships, we conduct a language-guided object retrieval experiment employing diverse query types across three semantic complexity levels (\textbf{Descriptive}: E.g., ``These are some books and they are on the table."; \textbf{Affordance}: E.g., ``Something I can open with my keys."; \textbf{Negation}: E.g., ``Something to sit on other than a chair.") provided by ConceptGraphs \cite{gu2023conceptgraphsopenvocabulary3dscene} on the Replica dataset \cite{replica19arxiv}. Following ConceptGraphs, we select 20 Descriptive, 5 Affordance and Negation queries for each scene, ensuring each query corresponds to at least one ground-truth object.

We employ three distinct object retrieval methods: (1) Semantic-based match: Objects in the scene graph are selected based on the cosine similarity between their semantic features and the CLIP text embedding of the query. The object exhibiting the highest similarity is chosen. (2) LLM-based match: LLM (GPT-4o) is utilized to identify the object node within the scene graph that optimally corresponds to the query statement. (3) Hierarchical scene graph-based match: For descriptive queries incorporating inter-object relationships, DynamicGSG leverages multi-layer scene graphs and object semantic features to perform hierarchical matching.


Analysis of top-1, top-2 and top-3 recall across diverse query types in Tab. \ref{tab:objret} indicates that hierarchical matching substantially enhances accuracy for descriptive queries involving inter-object relationships, exhibiting a significant improvement over ConceptGraphs \cite{gu2023conceptgraphsopenvocabulary3dscene}. This performance improvement is qualitatively illustrated in Fig. \ref{fig:obj_retrieval},
where our hierarchical matching approach successfully localizes target objects that baseline fails to identify. Futhermore, LLMs demonstrate superior instruction comprehension for Affordance and Negation queries, which more closely approximate natural human language. DynamicGSG also facilitates earlier object localization within these queries. 

\subsection{Scene Reconstruction Quality}
\label{subsec:reconstruction_quality}

To evaluate reconstruction fidelity of DynamicGSG relative to recent scene graph construction methods, we establish an evaluation protocol across three metrics (PSNR, SSIM, and LPIPS) following SplaTAM \cite{keetha2024splatam}. For a fair comparison, all methods utilize the ground-truth camera poses provided by the datasets. 


The quantitative comparisons on the Replica dataset \cite{replica19arxiv}, presented in Table \ref{tab:replica}, demonstrate that our method significantly outperforms point cloud based methods \cite{gu2023conceptgraphsopenvocabulary3dscene,werby23hovsg} while achieving comparable results to the advanced Gaussian semantic SLAM method SGS-SLAM \cite{Li_2024}, which utilizes ground-truth semantic annotations during optimization. To further validate the reconstruction quality of DynamicGSG, we extend our evaluation to novel view synthesis with SplaTAM on two scenes (8b5caf3398, b20a261fdf) from the ScanNet++ dataset \cite{yeshwanthliu2023scannetpp}. The results in Tab. \ref{tab:scannetpp} show that our method marginally outperforms SplaTAM in challenging scenarios.


\subsection{Real-world Dynamic Update}
\label{subsec:real_world}


\begin{table}[!t]
    \centering
    \begin{tabular}{lcc}\toprule
\multirow{1}{*}{\textbf{Types of Change}}      &            & \textbf{Success Rate (\%)}  \\ \midrule 

Object Disappearance      & 27 / 30              & 90.0  \\

Object Relocation & 25 / 30                     & 83.3      \\

Novel Object Emergence  & 19 / 20             & 95.0       \\
\midrule
\textbf{Total}   & 71 / 80                  & 88.8              \\
 \bottomrule
\end{tabular}
    \caption{\textbf{Success Rate of Dynamic Updates in Real-world.}}
    \label{tab:changes}
    \vspace{-2em}
\end{table}

To assess DynamicGSG's capability in adapting to dynamic environments, we establish 30 scenes in our lab with a total of 80 manual environment modifications, including object disappearance (30 instances), object relocation (30 instances), and novel object appearance (20 instances) and employ VINS-Fusion \cite{qin2019a} integrated with an Intel RealSense D455 to acquire aligned RGB-D streams at a resolution of 640×480, along with initial pose estimation.

As detailed in Tab. \ref{tab:changes}, DynamicGSG, while leveraging the method detailed in Sec. \ref{para:dynamic_update} to perform initial pose refinement and incrementally construct Gaussian scene graphs, successfully detects environment changes and executes corresponding dynamic updates, ensuring
temporal consistency between the scene graphs and the real-world environments. Some visualization results of the experiment, presented in Fig. \ref{fig:teaser} and \ref{fig:dynamic}, demonstrate DynamicGSG detects three types of environment changes and effectively performs instance-level local updates to construct dynamic, high-fidelity Gaussian scene graphs.

\section{Conclusion}
\label{sec:conclusion}
In this paper, we introduce DynamicGSG, a novel system designed to construct dynamic high-quality 3D Gaussian scene graphs. Utilizing fast differentiable rendering of 3D Gaussians, our system alleviates key problems in 3D scene graphs, such as the absence of mechanisms for dynamic environment adaptation and poor reconstruction quality. Extensive experimental results demonstrate that our system can perform dynamic updates in scene graphs according to real environment changes, effectively represent the spatial and semantic relationships between objects, and accurately capture intricate geometric details of scenes. These capabilities enable our system to assist agents in performing long-term navigation and mobile manipulation within indoor environments.



\bibliographystyle{IEEEtran}

\end{document}